\documentclass{article}

\usepackage{nicefrac}       
\usepackage{booktabs}      
\usepackage{subfigure}
\usepackage{natbib}
\usepackage{hyperref}
\usepackage{graphicx}
\usepackage{amsmath, amsthm, amsfonts}

\begin{document}

\title{An interpretable multiple kernel learning approach for the discovery of integrative cancer subtypes}
\author{Nora K. Speicher\,$^{1}$ and Nico Pfeifer\,$^{1,2,}$ \footnote{to whom correspondence should be addressed}\\
\fontsize{9pt}{1}\selectfont $^{1}$Department of Computational Biology and Applied Algorithmics, Max Planck\\
\fontsize{9pt}{1}\selectfont  Institute for Informatics, Saarland Informatics Campus Saarbr\"ucken, Germany\\
\fontsize{9pt}{1}\selectfont $^{2}$Methods in Medical Informatics,\\
\fontsize{9pt}{1}\selectfont Department of Computer Science, University of T\"ubingen, Germany}
\date{}

\maketitle
\textbf{Abstract:}
Due to the complexity of cancer, clustering algorithms have been used to disentangle the observed heterogeneity and identify cancer subtypes that can be treated specifically. While kernel based clustering approaches allow the use of more than one input matrix, which is an important factor when considering a multidimensional/manifold disease like cancer, the clustering results remain hard to evaluate and, in many cases, it is unclear which piece of information had which impact on the final result. In this paper, we propose an extension of multiple kernel learning clustering that enables the characterization of each identified patient cluster based on the features that had the highest impact on the result. To this end, we combine feature clustering with multiple kernel dimensionality reduction and introduce FIPPA, a score which measures the feature cluster impact on a patient cluster.\\
\textbf{Results:} We applied the approach to different cancer types described by four different data types with the aim of identifying integrative patient subtypes and understanding which features were most important for their identification. Our results show that our method does not only have state-of-the-art performance according to standard measures (e.g., survival analysis), but, based on the high impact features, it also produces meaningful explanations for the molecular bases of the subtypes. This could provide an important step in the validation of potential cancer subtypes and enable the formulation of new hypotheses concerning individual patient groups. Similar analysis are possible for other disease phenotypes.\\
\textbf{Availability:} Source code for rMKL-LPP with feature clustering and the calculation of fFIPPA scores is available at \href{https://github.molgen.mpg.de/nora/fFIPPA}{https://github.molgen.mpg.de/nora/fFIPPA}.\\
\textbf{Contact:} \href{pfeifer@informatik.uni-tuebingen.de}{pfeifer@informatik.uni-tuebingen.de}

\section{Introduction}
Despite the growing amount of multidimensional data available for cancer patients, the optimal treatment for a new patient remains in many cases unknown. In this setting, clustering algorithms are increasingly helpful to identify cancer subtypes, i.e., homogeneous groups of patients that could benefit from the same treatment.
Most clustering algorithms, e.g., k-means clustering~\citep{Hartigan:1975} or spectral clustering~\citep{Luxburg:2007} generate a hard clustering, i.e., each sample is assigned to one cluster, which works well when the clusters are separated. When analyzing cancer data, samples might lie between two or more clusters because they show characteristics of each of them. In these cases, fuzzy or soft clustering methods give additional information, since they assign to each patient a membership probability for each cluster instead of a binary cluster assignment. 
For the purpose of soft-clustering, fuzzy c-means~\citep{Dunn:1974, Bezdek:1981} extends the prominent k-means clustering method. To handle the increasing diversity in the data describing the same sample, fuzzy c-means can be combined with multiple kernel learning approaches, either directly as shown by~\citet{Hsin-ChienHuang:2012} or via application to the result of multiple kernel based dimensionality reduction~\citep{Speicher:2015}.

However, after the process of calculating kernel matrices for the input data, the subsequent data integration and the clustering of the samples, interpreting the results is difficult and the crucial features usually remain obscure. This lack of interpretability also hinders the validation of the identified groups in cases where the ground truth is not known.

We base our approach on the assumption that sample clusters are not defined by all features but rather by a subset of features which is also the motivation for the technique of biclustering or co-clustering, i.e., the simultaneous clustering of samples and features~\citep{Hartigan:1972}. This assumption becomes even more realistic with the increasing number of features within the data types, e.g., for DNA methylation or gene expression data, that are being measured nowadays for a cancer sample. Therefore, our approach combines feature clustering with subsequent data integration and sample clustering, thereby providing several advantages. First, the feature clustering identifies groups of homogeneous features, as typically, even for patients in the same cluster, not all measured features will behave similarly. Consequently, the sample similarities obtained from each of these feature groups are less noisy than the similarities one would obtain based on all the features. Second, due to the learned kernel weights we are able to identify which feature clusters were especially influential for each sample clustering allowing the further characterization of the identified sample clusters.

\paragraph{Related work }
Data integration has been combined with clustering algorithms in different ways. One approach that is regularly used for cancer data is the so-called \textit{late integration}, where clusterings are combined after they have been retrieved for each data source independently~\citep{TCGA:2012}. However, signals that might be spread over distinct data types could be too weak in each data source to have an impact on the individual clustering and therefore, might not be present in the final result.
To circumvent this problem, different approaches integrate the data at an earlier time point, this way performing \textit{intermediate integration}.

\textit{iCluster} creates the clustering using a joint latent variable model in which sparsity of the features is imposed by using an $l_1$ norm or a variance-related penalty~\citep{Shen:2012}. The authors applied their approach to a dataset of 55 glioblastoma patients described by three data types. In a different approach, \citet{Wang:2014} extract the sample similarities from each data type and generate sample similarity networks. These can then be fused iteratively by using a message-passing algorithm until convergence to a common integrated network.

Another possibility for data integration is multiple kernel learning whose strength lies in its wide applicability in different scenarios, e.g., to integrate data from arbitrary sources or to integrate distinct representations of the same input data. Therefore, it has been combined with a number of unsupervised algorithms in the last years.
\citet{Yu:2012} developed a multiple kernel version of k-means clustering. The optimization of the weights for the conic sum of kernels is a non-convex problem tackled by a bi-level alternating minimization procedure. A different approach for multiview k-means combines non-linear dimensionality reduction based on multiple kernels with subsequent clustering of the samples in the projection space~\citep{Speicher:2015}. \citet{Kumar:2011} extend spectral clustering such that it allows the clustering based on multiple inputs. Here, data integration is achieved by co-regularization of the kernel matrices which enforces the cluster structures in the different matrices to agree. Moving one step further, \citet{Gonen:2014} introduced localized kernel k-means, which, instead of optimizing one weight per kernel, uses per-sample weights to have more flexibility to account for sample-specific characteristics.
Some of the algorithms described have been applied to cancer data and could show improvements, such as increased robustness to noise in individual data sources and increased flexibility of the model~\citep{Gonen:2014, Speicher:2015}. 
However, they do not allow the extraction of importances directly related to specific features or the interpretation of the result on the basis of the applied algorithm, which is especially in unsupervised learning an important step to be able to associate clusters with their specific characteristics.

A number of approaches aim at increasing performance and interpretability by incorporating prior knowledge. Multiview biclustering~\citep{LeVan:2016} integrates copy number variations with Boolean mutation data, which are transformed into ranked matrices using a known network of molecular interactions between the genes. The subtype identification problem is then formulated as a rank matrix factorization of the ranked data. On breast cancer data, the method could refine the established subtypes which are based on PAM50 classification.
Approaches similar to ours concerning the subdivision into feature groups before applying multiple kernel learning were presented by \citet{Sinnott:2018} for supervised survival prediction of cancer patients and by~\citet{Rahimi:2018} for discriminating early- from late-stage cancers, showing a better interpretability compared to standard approaches. However, in both cases the feature groups are identified using prior knowledge (i.e., gene groups reflecting the memberships to pre-defined biological pathways) and the known outcome of interest is used to train the models, which is not the case for unsupervised clustering.

\section{Approach}
Due to the flexibility of the kernel functions, methods that can handle multiple input kernel matrices allow the integration of very distinct data types, such as numerical, sequence, or graph data, as well as the integration of different preprocessings of the input data.  Additionally, the individual up- or downweighting of kernel matrices can account for differences in the quality or relevance of the kernel matrices. However, the features within one data type can be very heterogeneous and might not all be equally important to a meaningful distance measure between the patients. Moreover, for non-linear kernels in general, it is not straightforward how to identify feature importances related to the result. Here, we propose a procedure that combines feature clustering with sample clustering based on multiple kernels, thereby increasing the potential to interpret the result without losing the power of the multiple kernel learning approach.
\begin{figure*}[!htb]
  \includegraphics[width=\textwidth, page=1]{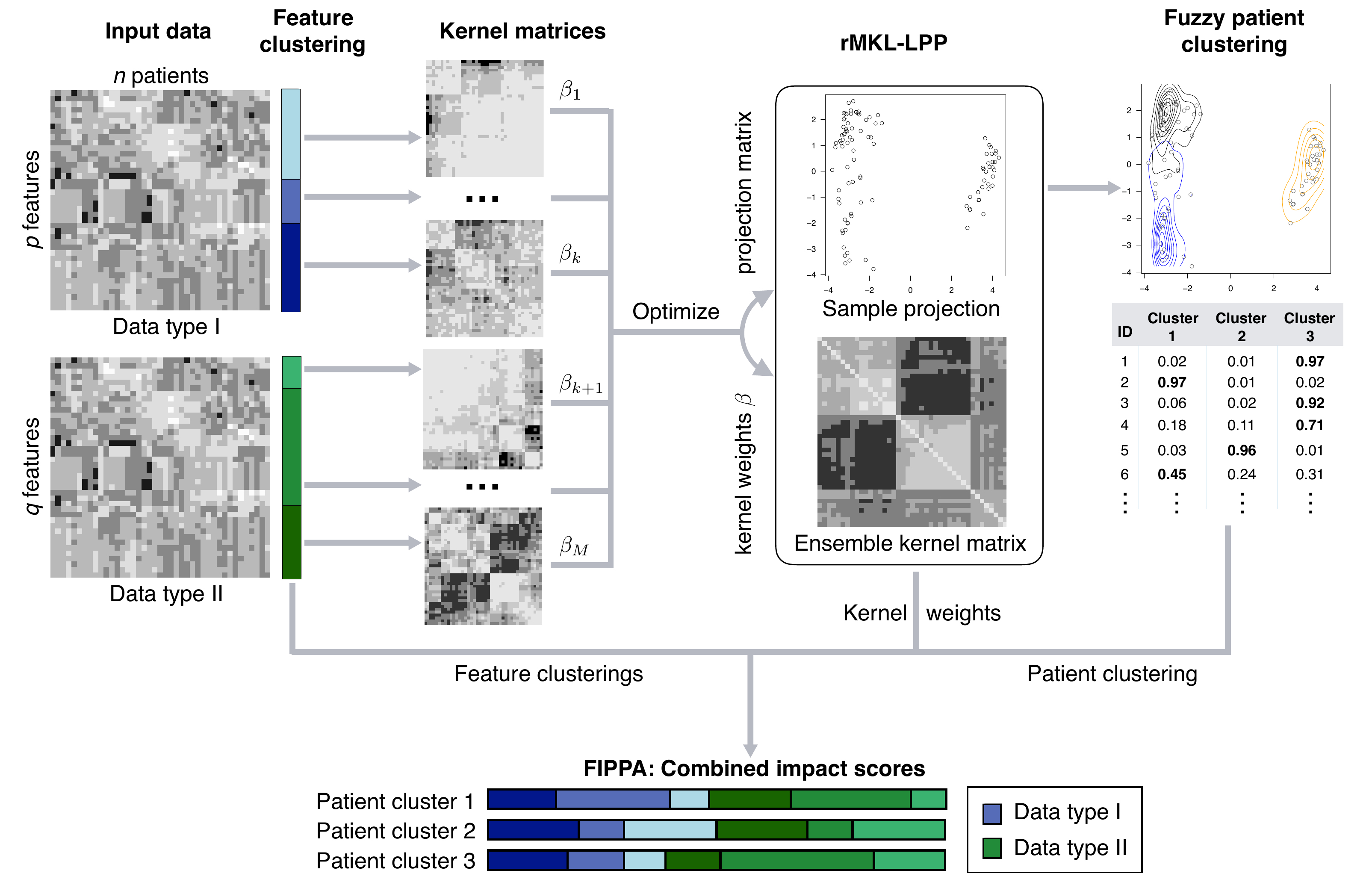}
  \caption{Overview of the presented approach for two input data matrices. First, feature clustering is performed, here with $C=3$. Each feature cluster gives rise to one kernel matrix, which are integrated using rMKL-LPP. This method optimizes one weight for each kernel matrix and a projection matrix, leading to a low-dimensional representation of the samples. Using these learned coordinates, the samples are clustered using fuzzy c-means. Finally, the feature clusters, kernel weights per feature cluster, and the patient clusters are used to calculate FIPPA scores, which describe feature cluster impact on a patient cluster.}
    \label{fig:overview}
\end{figure*}

As illustrated in Figure~\ref{fig:overview}, we cluster the features of each data type using k-means such that we can generate one kernel matrix based on each feature cluster. The kernel matrices are then integrated using a multiple kernel learning approach, here regularized multiple kernel learning for locality preserving projection. Based on the low-dimensional projection, we cluster the samples using fuzzy c-means. Our approach provides two advantages compared to standard procedures. First, increasing the homogeneity of the features by identifying feature clusters can reduce the noise in each kernel matrix, this way, a signal that is generated by only a few features can still significantly influence the final result if these features behave very similarly over a subset of samples. Second, the availability of the feature clusters and respective kernel weights allows for increased interpretability of the identified patient clusters as each cluster can be traced back to those groups of features that had the highest influence on the sample similarities.
To our knowledge, this is the first extension of a multiple kernel clustering algorithm towards integrative biclustering, where biclustering describes the simultaneous clustering of features and samples, or of rows and columns of a data matrix~\citep{Hartigan:1972}.

\section{Materials and methods}

\subsection{Regularized multiple kernel learning for locality preserving projections (rMKL-LPP)}
Given a set of $M$ kernel matrices $\{K_1, ..., K_M\}$ that are based on $N$ samples. Multiple kernel learning describes the optimization of a specific weight $\beta$ for each kernel matrix, such that an ensemble kernel matrix $ \mathbb{K}$ can be calculated via:
\begin{equation}
    \mathbb{K} = \sum_{m=1}^M \beta_m K_M, ~~\text{with}~~\beta_m \geq 0.
\end{equation}
Regularized multiple kernel learning for locality preserving projections \citep{Speicher:2015} optimizes these kernel weights together with a projection matrix such that local neighborhoods are preserved optimally when reducing the dimensionality of the data. The projection matrix $A \in \mathbb{R}^{p \times N}$ and the kernel weights $\boldsymbol{\beta} \in \mathbb{R}^M$ are learned via the minimization problem
\begin{align} 
	\underset{\boldsymbol{A}, \boldsymbol{\beta}}{\mbox{minimize }}~& \sum_{i,j=1}^N \| \boldsymbol{A}^T \mathbb{K}^{(i)} \boldsymbol{\beta} - \boldsymbol{A}^T \mathbb{K}^{(j)} \boldsymbol{\beta} \| ^2 w_{ij} \\
	\label{eq:rMKLDR} 
	\mbox{subject to }~& \sum_{i=1}^N \| \boldsymbol{A}^T \mathbb{K}^{(i)} \boldsymbol{\beta} \| ^2 d_{ii} = const.\\ 
&\| \boldsymbol{\beta} \|_1 = 1\\ 
   & \beta_m \geq 0, m = 1,2, ..., M. 
\end{align}
\noindent with
\begin{equation} 
\mathbb{K}^{(i)} = 
 \begin{pmatrix} 
  K_1 (1,i) & \cdots & K_M(1,i) \\ 
  \vdots   & \ddots & \vdots  \\ 
  K_1 (N,i) & \cdots & K_M(N,i) \\ 
 \end{pmatrix} \in \mathbb{R}^{N\times M}. 
\end{equation}

The projection matrix $A$ consists of $p$ vectors $\lbrack\boldsymbol{\alpha_1} \cdots \boldsymbol{\alpha_p} \rbrack$ that project the data into $p$ dimensions. The weight matrix $W$ and the diagonal matrix $D$ are determined by locality preserving projections using the $k$-neighborhood $\mathcal{N}_k(j)$ of a sample $j$ as follows
\begin{align}
    w_{ij} &= \begin{cases} 
                1,~~\text{if}~ i \in \mathcal{N}_k(j) \vee  j \in \mathcal{N}_k(i)\\ 
                0,~~\text{else} 
            \end{cases} \\
         \text{and }~~~~~ d_{ii} &= \sum_{i \neq j} w_{ij}.
\end{align}
The optimization of $A$ and $\beta$ is performed using coordinate descent. $A$ is optimized by solving a generalized eigenvalue problem, whereas the optimal $\beta$ is identified using semidefinite programming.

\subsection{Fuzzy c-means clustering (FCM)} Given a data matrix $X \in \mathbb{R}^{N\times l}$ describing $N$ samples $x_i$ with $l$ features, FCM identifies $C$ cluster centers $v_c$ and assigns the degrees of memberships $u_{i,c}$ for each sample $i$ and cluster $c$ by minimization of the following objective function~\citep{Dunn:1974, Bezdek:1981}:
\begin{align}
    \label{eq:fcm}
    J(U, v)~=~&\sum_{i=1}^N\sum_{c=1}^C u_{i,c}^f \lVert x_i- v_c \rVert ^2~~~~~~~~~~~~~~~~~~\\
    \text{subject to}~~ &\sum_{c=1}^C u_{i,c} =1~~~~~ \forall i\nonumber\\
    & u_{i,c} \geq 0 ~~~~~\forall i,c\nonumber\\
    & \sum_{i=1}^N u_{i,c} >0 ~~~~~\forall c. \nonumber
\end{align}
The resulting $U$ is an $N \times C$ matrix of the degrees of cluster memberships $u_{i,c}$, which depends on $f\geq1$, a parameter of the method that controls the degree of fuzzification. Choosing $f=1$ results in the hard clustering of the samples (i.e., $u_{i,c} \in \{0,1\}$); choosing $f \rightarrow \infty$ results in uniform cluster probabilities $u_{i,c}=\nicefrac{1}{C}$ for all $i,j$.

\subsection{Increased interpretability due to simultaneous clustering of features and samples}
\label{sect:impact}
The impact of each feature cluster $m \in \{1, .. , M\}$ on each identified sample cluster $c \in \{1, ..., C\}$ (FIPPA$_{c,m}$) can be calculated based on the kernel weights $\beta$ and the kernel matrices $K_m$ using the following equation
\begin{equation}
    \label{eq:imp}
    \text{FIPPA}_{c,m} = \dfrac{1}{|c|^2}\sum_{x_i, x_j \in c} \dfrac{\beta_m K_m[i,j]}{\mathbb{K}[i,j]},
\end{equation}
with $\mathbb{K}$ being the ensemble kernel matrix.
Here, we suppose a hard clustering of the patients, which can either be generated using a hard clustering algorithm or using the modal class of a fuzzy clustering.
However, fuzzy clustering provides additional information concerning the reliability of the cluster assignment for each sample. Using these probabilities can make the results more robust given that some samples might have an ambiguous signature and therefore lie between two or more clusters. Given a probability $p_c(x_i)=p(x_i \in c)$ that patient $x_i$ belongs to cluster $c$, the fuzzy FIPPA can be calculated as follows: 
\begin{equation}
    \label{eq:impFuzzy}
    \text{fFIPPA}_{c,m} = \dfrac{1}{|N|^2}\sum_{i, j=1}^N p_c(x_i)p_c(x_j) \dfrac{\beta_m K_m[i,j]}{\mathbb{K}[i,j]}.
\end{equation}
The incorporation of the joint probability $p(x_i\in c \land x_j\in c) = p_c(x_i)p_c(x_j)$ replaces the selection of sample pairs performed in Equation~\ref{eq:imp} but still allows to downweight the effects of samples that are uncertain or unlikely to belong to $c$.

Moreover, we can separate this overall score into a positive part ($ \text{fFIPPA}^+_{k,m}$) that leads to high intra-cluster similarity and a negative part that leads to high inter-cluster ($\text{fFIPPA}^-_{k,m}$) dissimilarity. For this purpose, we define
\begin{equation}
    \label{eq:Kplus}
    \mathbb{K}^+ = \sum_{m=1}^M \beta_m K_m^+~~~~ \text{and}~~~~
    \mathbb{K}^- = \sum_{m=1}^M \beta_m K_m^-
\end{equation}
with $K_m^+$ being the positive part of the matrix $K_m$ (all negative values set to zero) and vice versa for $K_m^-$. For this step, all kernel matrices need to be centered in the feature space, such that the mean of each matrix is equal to zero. When combining Formula~\ref{eq:Kplus} with Formula~\ref{eq:impFuzzy}, the calculation of the positive and negative fFIPPA is given by
\begin{align}
    \label{eq:impPosNeg}
    \text{fFIPPA}^+_{c,m} &= \dfrac{1}{|N|^2} \sum_{i, j = 1}^N p_c(x_i\land x_j) \dfrac{\beta_m K^+_m[i,j]}{\mathbb{K^+}[i,j]}, ~~~~\text{and}\nonumber\\
    \text{fFIPPA}^-_{c,m} &= \dfrac{1}{|N|^2}\sum_{i, j = 1}^N p_c(x_i \oplus x_j) \dfrac{\beta_m K^-_m[i,j]}{\mathbb{K^-}[i,j]}.
\end{align}

Besides the fact that the two scores are based on the positive and the negative part of the kernel matrices, respectively, the main difference between them is the probability factor for each summand. For the positive fFIPPA, the joint probability $p_c(x_i \land x_j)$ is used to generate high influence for pairs where both partners are likely to belong to cluster $c$, while for the negative fFIPPA, the \textit{exclusive or}, defined by
\begin{equation}
p_c(x_i \oplus x_j) = (p_c(x_i)+p_c(x_j)-2p_c(x_i)p_c(x_j)),
\end{equation}
results in an increased factor for pairs of samples of which exactly one has a high probability for $c$.

The fFIPPA scores calculated allow the identification of feature clusters that contribute more than average to the similarity of the samples within a sample cluster and the dissimilarity of the samples in two different clusters, thereby, revealing the underlying basis of the generated clustering.

\subsection{Materials}
We applied our approach to six different cancer data sets generated by The Cancer Genome Atlas which were downloaded from the UCSC Xena browser~\citep{Goldman:2017}. The cancer types covered are breast invasive carcinoma (BRCA), lung adenocarcinoma (LUAD), head and neck squamous cell carcinoma (HNSC), lower grade glioma (LGG), thyroid carcinoma (THCA) and prostate adenocarcinoma (PRAD). For each cancer patient, we used DNA methylation, gene expression data, copy number variations, and miRNA expression data for clustering. DNA methylation was mapped from methylation sites to gene promoter and gene body regions using RnBeads~\citep{Assenov:2014}. Using GeneTrail~\citep{Stoeckel:2016}, we mapped the miRNAs to their target genes (i.e., the gene that is regulated by the miRNA) such that the features of each data type were genes.
Due to the high number of features available, we performed the whole analysis on the 10\% of the features with the highest variance from each data type. For the cluster evaluation, we further leveraged the survival times of the patients for all cancer types except PRAD and THCA. For these two cancer types, survival analysis would not be informative due to the low number of events.

\section{Results and discussion}

\subsection{Parameter selection}
When applying our approach to a data set, we need to choose the number of feature clusters per data type as well as the number of patient clusters. For our experimental validation, we set both parameters to the same value ($c \in \{2, ..., 6\}$). Feature clustering was performed using k-means before generating the kernel matrices using the Gaussian radial basis kernel function. The parameter of the kernel $\gamma$ was chosen dependent on the number of features $d$ in the respective feature set based on the rule of thumb $\gamma=\frac{1}{2d^2}$~\citep{Gaertner:2002}. We generated three kernels per data type by multiplying this $\gamma$ with a factor $f_{\gamma} \in \{0.5, 1, 2\}$ and only used the one kernel matrix providing the highest variance in the first $d$ principal components.
 The number of neighbors for rMKL-LPP was set to 9 and the the number of dimensions $d$ to 5, as explained in our previous work~\citep{Speicher:2015}. 
The fuzzification degree $f$ of the soft-clustering algorithm was set to the default value of $2$ in concordance with~\citet{Dunn:1974}. If necessary for the subsequent analysis, we assigned each patient to its modal cluster (i.e., the cluster with the highest probability), otherwise, we used the cluster membership probabilities returned by FCM.

\subsection{Robustness}
\begin{figure}
  \centerline{\includegraphics[width=0.75\columnwidth]{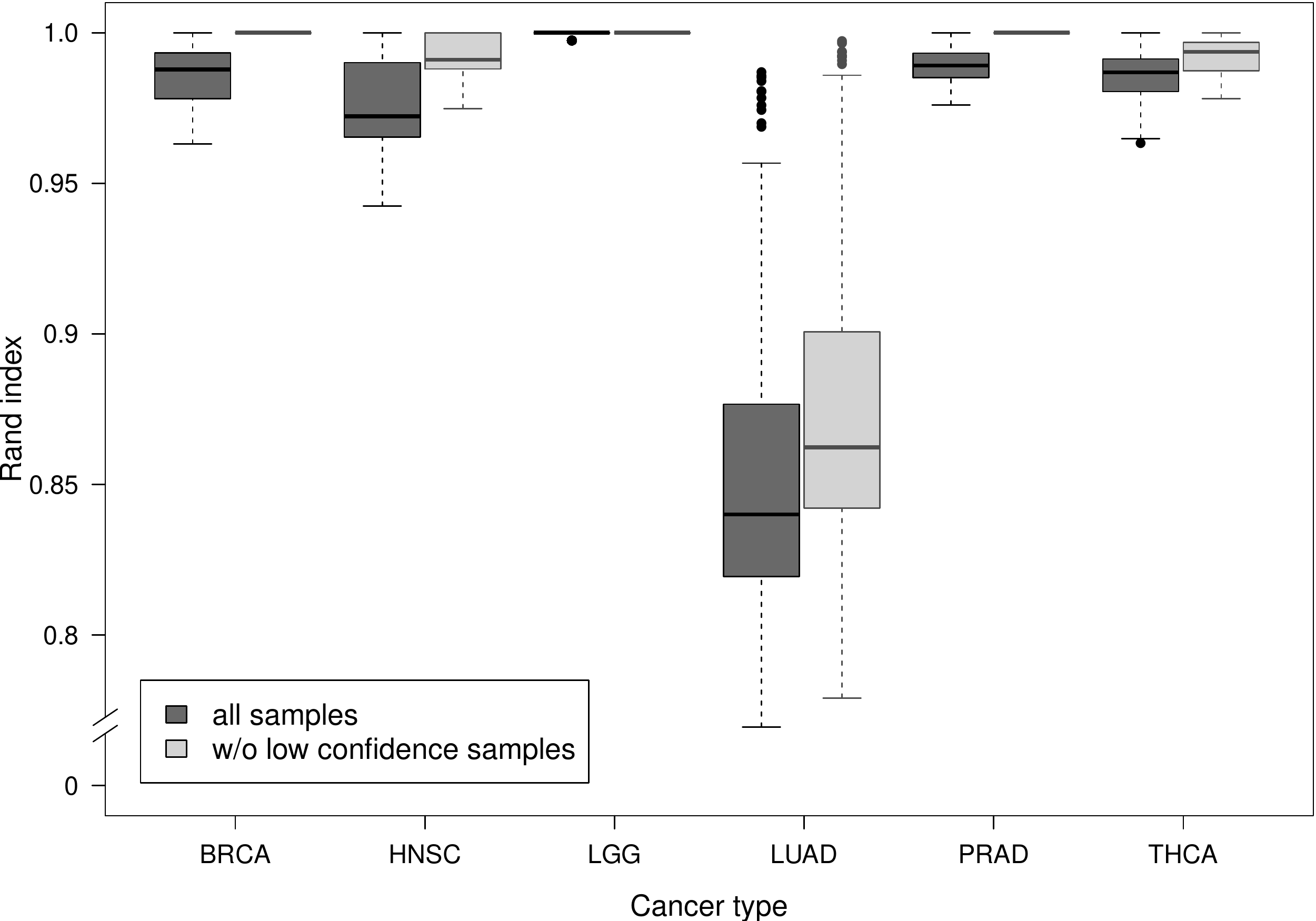}}
  \caption{Robustness of FC+rMKL-LPP in 50 repetitions. Dark grey boxes indicate Rand indices on the basis of all patients, light grey boxes indicate Rand indices calculated without low confidence samples, i.e., patients with a modal class probability $p_c$ more than one standard deviation smaller than the mean of the modal class probabilities.}
  \label{fig:robustnessBiCls}
\end{figure}

Since each kernel matrix in this approach is constructed on the basis of a feature cluster, slight changes in these initial clusters will propagate through the algorithm and might have an effect on the final result. We repeated the complete approach 50 times with different seeds to analyze the robustness of the final patient clustering. Figure~\ref{fig:robustnessBiCls} shows the pairwise similarities measured by the Rand index between the final patient clusterings. When including all patients according to their modal class, we observe high reproducibility for all cancer types except LUAD, for which the Rand index is approx. 0.85. When using the class probabilities for each sample to exclude patients where the prediction has a low confidence, here defined as being more than one standard deviation lower than the mean, we observe that the cluster assignments for the remaining patients are more stable for all cancer types including LUAD.

\begin{table*}[!ht]
  \caption{For each cancer type and method, we report the most significant results when varying the number of clusters from two to six. Average kLPP stands for kernel locality preserving projection on the (unweighted) average kernel, rMKL-LPP for the standard multiple kernel learning approach with one kernel per data type, and FC~+~rMKL-LPP represents the proposed approach for which the reported p-values are the median of 50 runs. All three approaches are combined with subsequent fuzzy c-means clustering. $C$ indicates the chosen number of clusters.}
  \label{tab:surv}
  \centering
  \begin{tabular}{p{1.5cm}|p{1.7cm}p{0.8cm}|p{1.7cm}p{0.5cm}|p{1.7cm} p{0.5cm}}
    \toprule
        Cancer  &\multicolumn{2}{c|}{average kLPP}&\multicolumn{2}{c|}{rMKL-LPP} & \multicolumn{2}{c}{FC + rMKL-LPP}\\
        & p-value & $C$ & p-value & $C$ &p-value & $C$\\
    \midrule
    BRCA &3.7E-2&6&  7.3E-2 & 6  & 5.0E-2 & 4 \\
    HNSC & 1.4E-3 &6&1.4E-3 & 6 & 9.96E-3 & 5  \\
    LGG  & $<$1.0E-16 & 3:6& $<$1.0E-16 & 3:6&$<$1.0E-16 & 3:6\\
    LUAD &0.15 &2&  2.9E-2 & 2 & 3.1E-2 & 6 \\
    \bottomrule
  \end{tabular}
\end{table*}

\subsection{Survival analysis}
To evaluate if a patient clustering could be clinically relevant, we performed survival analysis on the results with $C \in \{2, ..., 6\}$ clusters for each cancer type. To assess the influence of the feature clustering step, we also generated patient clusterings with two established methods: kLPP with the average kernel, and rMKL-LPP without feature clustering, which has been shown to perform well in comparison to other data integration methods \citep{Rappoport:2018}. To obtain a fair comparison, the same kernel matrices have been used for all approaches.

The p-value for the log-rank test~\citep{Hosmer:2011} indicates if there is at least one group of patients among the identified ones with a significant difference in the survival rate compared to the other groups. While the number of clusters is reflected in the degrees of freedom that are used in the test, we still corrected for the number of tests performed due to the variation of $C$ using Benjamini-Hochberg correction.
The results of the survival analysis shown in Table~\ref{tab:surv} indicate that on the four cancer types where the number of events allows for a survival analysis, our approach has a comparable performance to the established methods and is able to find biologically meaningful groups of patients. Moreover, despite the additional flexibility which is achieved by the feature clustering, the optimal number of clusters does not increase in comparison to less complex methods. However, as not all patients receive an optimal treatment, the survival data can be biased. Therefore, we need additional means to evaluate and interpret the underlying structure of the identified clusterings.

\subsection{Interpretation}
\begin{figure*}
  \centering
  \begin{subfigure}[fFIPPA scores for intra-cluster similarity of the patient clusters.]{
  \includegraphics[width=\textwidth, page=1]{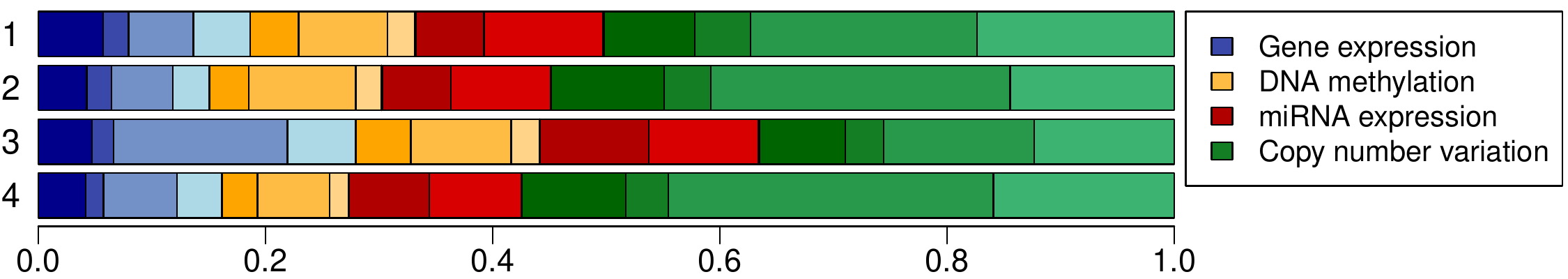}
     \label{fig:impPos}}
    \end{subfigure}
    \begin{subfigure}[fFIPPA scores for inter-cluster dissimilarity of the patient clusters.]{
  \includegraphics[width=\textwidth, page=2]{impact_BRCA.pdf}
     \label{fig:impNeg}}
    \end{subfigure}
    \caption{\textbf{Positive and negative fFIPPA of each feature cluster and patients cluster} for BRCA with $C=4$. Each row represents one patient cluster, i.e., it is derived using the probabilities for this specific cluster. The width of a bar segment is defined by $\text{fFIPPA}_{c,m}^+$ or ($\text{fFIPPA}_{c,m}^-$) for the respective patient cluster $c$ and feature cluster $m$.}
\end{figure*}

As described in Section~\ref{sect:impact} the weights for each feature cluster in combination with the kernel matrices allow us to calculate the fFIPPA scores. Figure~\ref{fig:impPos} visualizes the positive fFIPPA scores for the BRCA clustering with $c=4$. Here, we can see clear differences between the patient clusters. While for all of them, copy number variations have the strongest impact, the exact contributions of the data types vary (e.g. the fFIPPA of gene expression, especially feature cluster 3, is strongest in patient cluster 3). When comparing to the negative fFIPPAs (cf. Figure~\ref{fig:impNeg}), we observe in general similar patterns. Differences still occur, e.g., in patient cluster 2 and 4, one can see that the fFIPPA of copy number variations on the intra-cluster similarity is larger than their impact on inter-cluster dissimilarity.

Observing the impact of each data type is also possible with traditional multiple kernel learning, however, in our approach, we can additionally analyze the impact of the individual feature clusters.
To gain further insights into the underlying biological mechanisms, we used the positive and negative fFIPPA scores to identify which feature clusters contributed more than average to a patient cluster, i.e., we initialized two lists of feature clusters, the first one explaining high dissimilarity to other patient clusters, the second one explaining high similarity within the patient cluster. After identification of these high-fFIPPA feature clusters, we filtered the associated features to keep only those that behave homogeneously over the patients of the specific cluster. For the homogeneity filtering, we introduced a homogeneity score, counting in how many patients of the respective patient cluster the gene was constantly expressed (i.e., either over- or underexpressed, whichever happened more often). Based on these scores, we obtained a distribution representing the homogeneity of the features within a cluster. For further analysis, we kept every gene with a score larger than the mean of this distribution and distinguish between over- and underexpressed genes. This way, we generated four lists of features for each patient cluster describing a) overexpressed genes leading to high intra-cluster similarity, b) underexpressed genes leading to high intra-cluster similarity, c) overexpressed genes leading to high inter-cluster dissimilarity and d) underexpressed genes leading to high inter-cluster dissimilarity.

In these lists, we found genes with known implications to cancer formation or progression. Some of the genes were related to several cancer types (e.g. PTEN, which regulates the AKT/PKB signaling pathway and thereby represents a tumor suppressor in various tumors~\citep{Cantley:1999} was present in lists for BRCA, THCA, and PRAD) or to several clusters of the same cancer type (e.g.SOX2 for HNSC, which has been shown to control tumor initiation in squamous-cell carcinomas~\citep{Boumahdi:2014}). Others were only found for one specific cluster of a cancer type (e.g., among the four clusters of BRCA, only one gene list included GRPR, a G-protein coupled receptor that was shown to influence the viability of breast cancer cells~\citep{Cornelio:2013}). From a biological perspective, this finding is consistent with our expectation that the lists are not disjoint, since cancer does not form  due to an individual mutation but rather as a consequence of a number of different aberrations, which can be shared between different subtypes or even cancer types.

Besides looking at the lists at the level of individual genes, we also inferred biological functions that are related to these gene lists using GeneTrail~\citep{Stoeckel:2016}. We applied over-representation analysis which tests, if significantly more genes in the list are associated to a specific category than would be expected based on a background gene list. As background all the genes were used that entered the clustering and we tested for enrichment of KEGG pathways and gene ontology categories (GO).

As there was some variation in the cluster assignments, one would also expect some variation in the enriched terms. To quantify the stability of the result, we compared the identified terms of the best to the terms of the median result, where best and median was defined according to the survival p-value. The similarity of the GO terms was determined using the relevance score~\cite{Schlicker:2006}, for the similarity between two sets of KEGG pathways we used the Jaccard index. The average similarities correlate to the observed stabilities of the cluster assignments: The value for LUAD is slightly lower than for BRCA and HNSC (0.48 vs. 0.58 and 0.68, respectively), however, for all three cancer types the scores indicate that similar functions or pathways are found. For LGG, the results were stable such that the optimum and median were equivalent, for PRAD and THCA, the best and median result could not be identified due to the lack of adequate survival data.

Based on the intersection of the optimal and median result, Table~\ref{tab:HNSC_ORA} shows a subset of the findings for HNSC for Cluster 3, which has the poorest prognosis (median survival time 1079 days), and Cluster 4, which has the best prognosis (median survival time $>4241$ days). The terms identified are very different for the two clusters. In Cluster 3, we obtained enriched terms that are mainly associated with muscle cells, more specifically, their development or their activation. Skeletal muscle invasion seems to be correlated to the recurrence of the disease~\citep{Chandler:2011}, which agrees with our observation that patients having a distinct signature in the genes involved in these functions have a poor survival.
\begin{table}
  \caption{Significant GO terms and KEGG pathways associated with high intra-cluster similarity identified by over-representation analysis for patient clusters 3 and 4 for HNSC. For patient cluster 3, the terms are based on inactive genes, for patient cluster 4 on active genes.}
  \label{tab:HNSC_ORA}
  \centering
  \begin{tabular}{lp{10cm}}
    \toprule
       $c=3$&GO-Cellular component (p-value $<1.0E-3$)\\
    \midrule
&contractile fiber\\
 &   myofibril\\
  &  sarcomere\\
   & I band\\
    &myosin II complex\\
    &muscle myosin complex\\
    &A band\\
    &myosin complex\\
    &Z disc\\
    &myofilament\\
    \midrule
    &GO - Biological Process (p-value $<1.0E-3$)\\
\midrule
&myofibril assembly\\
    &striated muscle cell development\\
    &muscle filament sliding\&actin myosin filament sliding\\
    &striated muscle contraction\\
    &actomyosin structure organization\\
    &muscle cell development\\
     \midrule\midrule
   $c=4$& GO -Biological process (p-value $<1.0E-3$)\\
\midrule  
&regulation of peptidyl serine phosphorylation of STAT protein\\
&natural killer cell activation involved in immune response\\
&serine phosphorylation of STAT protein\\
&response to exogenous dsRNA\\
&natural killer cell activation\\
&positive regulation of peptidyl serine phosphorylation\\
&regulation of type I interferon mediated signaling pathway\\
      \bottomrule
  \end{tabular}
\end{table}
For patient cluster 4, we identified terms related to the \textit{phosphorylation of a STAT protein} which regulates as a part of the JAK/STAT signaling pathway cell growth and has been shown to play a role in tumor formation and progression~\citep{Song:2000}. On the other hand, the identified terms mainly refer to functions of the immune system, including the \textit{regulation of type I interferon mediated signaling pathway} which is a regulation of the innate immune response. Despite of the controversial role of inflammation and the immune system in cancer~\citep{Coussens:2002}, its activation could help eliminate cancer cells and therefore, be beneficial for the respective patient group.
For HNSC, the terms associated to high inter-cluster dissimilarity and high intra-cluster similarity differ only for Cluster 1, where we identify terms, such as \textit{B cell differentiation} and \textit{proliferation}, that are associated to high dissimilarity to other clusters but not to high intra-cluster similarity.

We also applied our approach to PRAD and THCA where survival analysis is not meaningful because only a small number of patients die from the disease. For both cancer types, we set $C$ to 4 and could identify different significant terms for some patient clusters. For PRAD, we identified amongst others a number of terms related to the activity of olfactory receptors, which have been shown to participate in the process of tumor cell proliferation and apoptosis~\citep{Chen:2018}. Several hits for THCA, e.g., \textit{DNA deamination}, indicate an influence of epigenetic regulation mechanisms, which is supported by current literature~\citep{Rodriguez-Rodero:2017}.

\section{Conclusion}

The approach presented in this paper provides a step towards a better interpretabilty of multiple kernel clustering. For a few kernel functions it is possible to extract feature importances, however, our new procedure is independent of the kernel function used and can consequently be applied in combination with any kernel function. The application to biological data sets shows that the approach gives state-of-the-art performance with respect to survival analysis. Additionally, the weights for the feature clusters can be used to extract meaningful information describing the individual sample clusters identified, thereby, leading to comprehensive insights on individual subtypes which allows to generate hypotheses that can be specifically tested. The initial screening for overlap between the found information and biological literature shows very promising results.

In our current approach, the feature clusters influence the patient clusters but not vice versa. As shown on biological examples, these feature groups already provide valuable information. However, to obtain feature clusters, that are even more specific to the patient clusters, simultaneous optimization of feature and patient clusters could be beneficial. Also are the feature clusters currently restricted to be non-overlapping and to include every feature. In many real-world scenarios we might want to a more flexible approach that is relaxing these constraints.

\bibliographystyle{natbib}
\bibliography{arxiv}{}

\begin{thebibliography}{}

\bibitem[Assenov {\em et~al.}(2014)Assenov, M{\"{u}}ller, Lutsik, Walter,
  Lengauer, and Bock]{Assenov:2014}
Assenov, Y., M{\"{u}}ller, F., Lutsik, P., Walter, J., Lengauer, T., and Bock,
  C. (2014).
\newblock {Comprehensive analysis of DNA methylation data with RnBeads}.
\newblock {\em Nature Methods\/}, {\bf 11}(11), 1138--1140.

\bibitem[Bezdek(1981)Bezdek]{Bezdek:1981}
Bezdek, J.~C. (1981).
\newblock {\em {Pattern Recognition with Fuzzy Objective Function
  Algorithms}\/}.
\newblock Plenum Press.

\bibitem[Boumahdi {\em et~al.}(2014)Boumahdi, Driessens, Lapouge, Rorive,
  Nassar, {Le Mercier}, Delatte, Caauwe, Lenglez, Nkusi, Broh{\'{e}}e, Salmon,
  Dubois, del Marmol, Fuks, Beck, and Blanpain]{Boumahdi:2014}
Boumahdi, S., Driessens, G., Lapouge, G., Rorive, S., Nassar, D., {Le Mercier},
  M., Delatte, B., Caauwe, A., Lenglez, S., Nkusi, E., Broh{\'{e}}e, S.,
  Salmon, I., Dubois, C., del Marmol, V., Fuks, F., Beck, B., and Blanpain, C.
  (2014).
\newblock {SOX2 controls tumour initiation and cancer stem-cell functions in
  squamous-cell carcinoma}.
\newblock {\em Nature\/}, {\bf 511}(7508), 246--250.

\bibitem[Cantley and Neel(1999)Cantley and Neel]{Cantley:1999}
Cantley, L.~C. and Neel, B.~G. (1999).
\newblock {New insights into tumor suppression: PTEN suppresses tumor formation
  by restraining the phosphoinositide 3-kinase/AKT pathway.}
\newblock {\em Proceedings of the National Academy of Sciences of the United
  States of America\/}, {\bf 96}(8), 4240--5.

\bibitem[Chandler {\em et~al.}(2011)Chandler, Vance, Budnick, and
  Muller]{Chandler:2011}
Chandler, K., Vance, C., Budnick, S., and Muller, S. (2011).
\newblock {Muscle invasion in oral tongue squamous cell carcinoma as a
  predictor of nodal status and local recurrence: just as effective as depth of
  invasion?}
\newblock {\em Head and neck pathology\/}, {\bf 5}(4), 359--63.

\bibitem[Chen {\em et~al.}(2018)Chen, Zhao, Fu, and Chen]{Chen:2018}
Chen, Z., Zhao, H., Fu, N., and Chen, L. (2018).
\newblock {The diversified function and potential therapy of ectopic olfactory
  receptors in non-olfactory tissues}.
\newblock {\em Journal of Cellular Physiology\/}, {\bf 233}(3), 2104--2115.

\bibitem[Cornelio {\em et~al.}(2013)Cornelio, {DE Farias}, Prusch, Heinen, {Dos
  Santos}, Abujamra, Schwartsmann, and Roesler]{Cornelio:2013}
Cornelio, D.~B., {DE Farias}, C.~B., Prusch, D.~S., Heinen, T.~E., {Dos
  Santos}, R.~P., Abujamra, A.~L., Schwartsmann, G., and Roesler, R. (2013).
\newblock {Influence of GRPR and BDNF/TrkB signaling on the viability of breast
  and gynecologic cancer cells.}
\newblock {\em Molecular and clinical oncology\/}, {\bf 1}(1), 148--152.

\bibitem[Coussens and Werb(2002)Coussens and Werb]{Coussens:2002}
Coussens, L.~M. and Werb, Z. (2002).
\newblock {Inflammation and cancer.}
\newblock {\em Nature\/}, {\bf 420}(6917), 860--7.

\bibitem[Dunn(1974)Dunn]{Dunn:1974}
Dunn, J.~C. (1974).
\newblock {A Fuzzy Relative of the ISODATA Process and Its Use in Detecting
  Compact Well-Separated Clusters}.
\newblock {\em Journal of Cybernetics\/}, {\bf 3}(3), 32--57.

\bibitem[G\"{a}rtner {\em et~al.}(2002)G\"{a}rtner, Flach, Kowalczyk, and
  Smola]{Gaertner:2002}
G\"{a}rtner, T., Flach, P.~A., Kowalczyk, A., and Smola, A.~J. (2002).
\newblock Multi-instance kernels.
\newblock In {\em Proc. 19th International Conf. on Machine Learning\/}, pages
  179--186. Morgan Kaufmann.

\bibitem[Goldman {\em et~al.}(2017)Goldman, Craft, Zhu, and
  Haussler]{Goldman:2017}
Goldman, M., Craft, B., Zhu, J., and Haussler, D. (2017).
\newblock { The UCSC Xena system for cancer genomics data visualization and
  interpretation [abstract]}.
\newblock {\em Proceedings of the American Association for Cancer Research
  Annual Meeting 2017; Cancer Research\/}, {\bf 77}(13 Supplement), 2584--2584.
\newblock Available from {http://xena.ucsc.edu} [Accessed: 2017/10/24].

\bibitem[G{\"{o}}nen and Margolin(2014)G{\"{o}}nen and Margolin]{Gonen:2014}
G{\"{o}}nen, M. and Margolin, A.~A. (2014).
\newblock {Localized Data Fusion for Kernel k-Means Clustering with Application
  to Cancer Biology}.
\newblock In {\em Advances in Neural Information Processing Systems 27 (NIPS
  2014)\/}, pages 1305--1313.

\bibitem[Hartigan(1972)Hartigan]{Hartigan:1972}
Hartigan, J.~A. (1972).
\newblock {Direct Clustering of a Data Matrix}.
\newblock {\em Journal of the American Statistical Association\/}, {\bf
  67}(337), 123--129.

\bibitem[Hartigan(1975)Hartigan]{Hartigan:1975}
Hartigan, J.~A. (1975).
\newblock {\em {Clustering Algorithms}\/}.
\newblock John Wiley {\&} Sons.

\bibitem[{Hosmer Jr.} {\em et~al.}(2011){Hosmer Jr.}, Lemeshow, and
  May]{Hosmer:2011}
{Hosmer Jr.}, D.~W., Lemeshow, S., and May, S. (2011).
\newblock {\em Applied Survival Analysis: Regression Modeling of Time to Event
  Data\/}.
\newblock Wiley.

\bibitem[{Hsin-Chien Huang} {\em et~al.}(2012){Hsin-Chien Huang}, {Yung-Yu
  Chuang}, and {Chu-Song Chen}]{Hsin-ChienHuang:2012}
{Hsin-Chien Huang}, {Yung-Yu Chuang}, and {Chu-Song Chen} (2012).
\newblock {Multiple Kernel Fuzzy Clustering}.
\newblock {\em IEEE Transactions on Fuzzy Systems\/}, {\bf 20}(1), 120--134.

\bibitem[Kumar {\em et~al.}(2011)Kumar, Rai, and Daume]{Kumar:2011}
Kumar, A., Rai, P., and Daume, H. (2011).
\newblock Co-regularized multi-view spectral clustering.
\newblock In J.~Shawe-Taylor, R.~S. Zemel, P.~L. Bartlett, F.~Pereira, and
  K.~Q. Weinberger, editors, {\em Advances in Neural Information Processing
  Systems 24\/}, pages 1413--1421. Curran Associates, Inc.

\bibitem[{Le Van} {\em et~al.}(2016){Le Van}, van Leeuwen, {Carolina Fierro},
  {De Maeyer}, {Van den Eynden}, Verbeke, {De Raedt}, Marchal, and
  Nijssen]{LeVan:2016}
{Le Van}, T., van Leeuwen, M., {Carolina Fierro}, A., {De Maeyer}, D., {Van den
  Eynden}, J., Verbeke, L., {De Raedt}, L., Marchal, K., and Nijssen, S.
  (2016).
\newblock {Simultaneous discovery of cancer subtypes and subtype features by
  molecular data integration}.
\newblock {\em Bioinformatics\/}, {\bf 32}(17), i445--i454.

\bibitem[Rahimi and G{\"{o}}nen(2018)Rahimi and G{\"{o}}nen]{Rahimi:2018}
Rahimi, A. and G{\"{o}}nen, M. (2018).
\newblock {Discriminating early- and late-stage cancers using multiple kernel
  learning on gene sets}.
\newblock {\em Bioinformatics\/}, {\bf 34}(13), i412--i421.

\bibitem[Rappoport and Shamir(2018)Rappoport and Shamir]{Rappoport:2018}
Rappoport, N. and Shamir, R. (2018).
\newblock {Multi-omic and multi-view clustering algorithms: review and cancer
  benchmark}.
\newblock {\em Nucleic Acids Research\/}.

\bibitem[Rodr{\'{i}}guez-Rodero {\em et~al.}(2017)Rodr{\'{i}}guez-Rodero,
  Delgado-{\'{A}}lvarez, D{\'{i}}az-Naya, {Mart{\'{i}}n Nieto}, and
  {Men{\'{e}}ndez Torre}]{Rodriguez-Rodero:2017}
Rodr{\'{i}}guez-Rodero, S., Delgado-{\'{A}}lvarez, E., D{\'{i}}az-Naya, L.,
  {Mart{\'{i}}n Nieto}, A., and {Men{\'{e}}ndez Torre}, E. (2017).
\newblock {Epigenetic modulators of thyroid cancer}.
\newblock {\em Endocrinolog{\'{i}}a, Diabetes y Nutrici{\'{o}}n\/}, {\bf
  64}(1), 44--56.

\bibitem[Schlicker {\em et~al.}(2006)Schlicker, Domingues, Rahnenf{\"{u}}hrer,
  and Lengauer]{Schlicker:2006}
Schlicker, A., Domingues, F.~S., Rahnenf{\"{u}}hrer, J., and Lengauer, T.
  (2006).
\newblock {A new measure for functional similarity of gene products based on
  Gene Ontology}.
\newblock {\em BMC Bioinformatics\/}, {\bf 7}(1), 302.

\bibitem[Shen {\em et~al.}(2012)Shen, Mo, Schultz, Seshan, Olshen, Huse,
  Ladanyi, and Sander]{Shen:2012}
Shen, R., Mo, Q., Schultz, N., Seshan, V.~E., Olshen, A.~B., Huse, J., Ladanyi,
  M., and Sander, C. (2012).
\newblock Integrative subtype discovery in glioblastoma using {iCluster}.
\newblock {\em PloS ONE\/}, {\bf 7}.

\bibitem[Sinnott and Cai(2018)Sinnott and Cai]{Sinnott:2018}
Sinnott, J.~A. and Cai, T. (2018).
\newblock {Pathway aggregation for survival prediction via multiple kernel
  learning}.
\newblock {\em Statistics in Medicine\/}.

\bibitem[Song and Grandis(2000)Song and Grandis]{Song:2000}
Song, J.~I. and Grandis, J.~R. (2000).
\newblock {STAT signaling in head and neck cancer}.
\newblock {\em Oncogene\/}, {\bf 19}(21), 2489--2495.

\bibitem[Speicher and Pfeifer(2015)Speicher and Pfeifer]{Speicher:2015}
Speicher, N.~K. and Pfeifer, N. (2015).
\newblock Integrating different data types by regularized unsupervised multiple
  kernel learning with application to cancer subtype discovery.
\newblock {\em Bioinformatics\/}, {\bf 31}(12), i268--i275.

\bibitem[St{\"{o}}ckel {\em et~al.}(2016)St{\"{o}}ckel, Kehl, Trampert,
  Schneider, Backes, Ludwig, Gerasch, Kaufmann, Gessler, Graf, Meese, Keller,
  and Lenhof]{Stoeckel:2016}
St{\"{o}}ckel, D., Kehl, T., Trampert, P., Schneider, L., Backes, C., Ludwig,
  N., Gerasch, A., Kaufmann, M., Gessler, M., Graf, N., Meese, E., Keller, A.,
  and Lenhof, H.-P. (2016).
\newblock {Multi-omics enrichment analysis using the GeneTrail2 web service}.
\newblock {\em Bioinformatics\/}, {\bf 32}(10), 1502--1508.

\bibitem[{The Cancer Genome Atlas}(2012){The Cancer Genome Atlas}]{TCGA:2012}
{The Cancer Genome Atlas} (2012).
\newblock {Comprehensive molecular portraits of human breast tumors}.
\newblock {\em Nature\/}, {\bf 490}, 61--70.

\bibitem[von Luxburg(2007)von Luxburg]{Luxburg:2007}
von Luxburg, U. (2007).
\newblock A tutorial on spectral clustering.
\newblock {\em Statistics and Computing\/}, {\bf 17}(4), 395--416.

\bibitem[Wang {\em et~al.}(2014)Wang, Mezlini, Demir, Fiume, Tu, Brudno,
  Haibe-Kains, and Goldenberg]{Wang:2014}
Wang, B., Mezlini, A.~M., Demir, F., Fiume, M., Tu, Z., Brudno, M.,
  Haibe-Kains, B., and Goldenberg, A. (2014).
\newblock Similarity network fusion for aggregating data types on a genomic
  scale.
\newblock {\em Nature Methods\/}, {\bf 11}(3), 333--337.

\bibitem[Yu {\em et~al.}(2012)Yu, Tranchevent, Liu, Glanzel, Suykens, {De
  Moor}, and Moreau]{Yu:2012}
Yu, S., Tranchevent, L.-C., Liu, X., Glanzel, W., Suykens, J. A.~K., {De Moor},
  B., and Moreau, Y. (2012).
\newblock {Optimized Data Fusion for Kernel k-Means Clustering}.
\newblock {\em IEEE Transactions on Pattern Analysis and Machine
  Intelligence\/}, {\bf 34}(5), 1031--1039.

\end{thebibliography}

\end{document}